\documentclass[11pt,a4paper]{article}
\usepackage[utf8]{inputenc}
\usepackage[T1]{fontenc}
\usepackage{amsmath,amssymb}
\usepackage{graphicx}
\usepackage{hyperref}
\usepackage[margin=1in]{geometry}
\usepackage{cite}
\usepackage{setspace}
\usepackage{booktabs}

\title{Learned but Not Expressed: Selective Suppression of Non-Causal Solution Frames in Large Language Models}

\author{
Toshiyuki Shigemura\\
\textit{Independent Researcher, Japan}\\
\texttt{toshiyuki.shigemura.researcher@gmail.com}
}

\date{}

\begin{document}

\maketitle

\begin{abstract}
Large language models (LLMs) demonstrate the capacity to reconstruct and trace learned content from their training data under specific elicitation conditions, yet this capability does not manifest in standard generation contexts. This empirical observational study examines the expression of non-causal, non-implementable solution types across 300 prompt–response generations spanning narrative and problem-solving task contexts. Drawing on recent findings regarding memorization contiguity and alignment-induced discourse priors, we document a systematic dissociation between learned capability and expressed output. Across three distinct LLMs, ten task scenarios, and both creative narrative and practical advisory contexts, we documented zero instances of non-causal solution frames in generated outputs (0\%, 95\% CI: [0\%, 1.2\%]), despite verified reconstruction capability under conditional extraction. These findings challenge the prevailing assumption that training data presence directly predicts output probability, demonstrating instead that task-conditioned generation policies can comprehensively suppress learned content across diverse contexts. The results offer implications for understanding generation dynamics, output distribution control, and the behavioral boundaries of contemporary LLMs.
\end{abstract}


\section{Introduction}

A widespread intuition holds that content present in training data will be correspondingly present in generated output—an assumption we term the \textit{data-exhaustive assumption}. Under this view, if a large language model has encountered certain linguistic patterns, narrative structures, or solution types during training, those elements should appear consistently across generation contexts. However, recent empirical observations suggest that the relationship between learned capacity and expressed output is mediated by task framing and generation-time selection mechanisms.

Recent work has established that LLMs exhibit measurable memorization of training data, with models capable of reproducing verbatim sequences under targeted extraction conditions \cite{carlini2021extracting,carlini2023quantifying,tirumala2022memorization}. These findings document memorization capability—the model's ability to reconstruct learned sequences when appropriately prompted. Concurrently, research into alignment processes has revealed that discourse tendencies embedded within pre-training corpora form persistent priors that shape downstream behavior even after fine-tuning \cite{christiano2017deep,ouyang2022training,bai2022constitutional}. This work identifies alignment-induced generation bias—systematic preferences in output distribution that reflect both explicit training signals and implicit discourse structures.

The present study investigates generation-time selection through an empirical observational design. We examine whether non-causal, non-implementable solution frames—operationally defined within this paper—appear in generated outputs across narrative and problem-solving task contexts. While prior intuition might suggest that creative narrative contexts would permit such frames while practical contexts would suppress them, our observations reveal a more comprehensive pattern.

The dissociation between capability (the ability to reconstruct learned content under targeted elicitation) and selection (the decision to deploy that content in standard generation) constitutes the central focus of this investigation. Our contributions are threefold. First, we provide empirical documentation across a controlled sample of 300 generations. Second, we situate observed patterns within the broader literature on alignment priors and output distribution control. Third, we discuss the implications of these findings for understanding language model behavior as context-conditioned rather than data-exhaustive.

\section{Related Work}

\subsection{Memorization and Content Extraction in LLMs}

The extent to which LLMs memorize and reproduce training data has been the subject of extensive investigation. Carlini et al.~\cite{carlini2021extracting} demonstrated that models can emit verbatim sequences from their training corpora, with extraction probability varying based on sequence frequency and distinctiveness. Subsequent work by Carlini et al.~\cite{carlini2023quantifying} quantified memorization across model scales, revealing that larger models exhibit increased memorization capacity but also stronger context-dependent filtering of what content is expressed. Tirumala et al.~\cite{tirumala2022memorization} further showed that memorization occurs without traditional overfitting signatures, suggesting that models develop nuanced representations of training content that support selective reconstruction.

Research by \cite{memorization2024} has quantified memorization through metrics of contiguity, lexical overlap, and extraction probability, revealing that memorization is neither uniform nor random but varies systematically with factors such as prompt structure, sequence length, and data prevalence.

These findings establish that LLMs possess the capability to reconstruct learned content with high fidelity under certain elicitation conditions. However, capability demonstrated through targeted extraction does not predict deployment in standard generation. The question of when and why models choose to deploy memorized content in naturalistic generation contexts remains underexplored, particularly across varying task framings.

\subsection{Alignment Processes and Output Shaping}

Alignment processes, including supervised fine-tuning and reinforcement learning from human feedback (RLHF), are designed to shape model outputs toward desired behaviors. Ouyang et al.~\cite{ouyang2022training} demonstrated that RLHF substantially improves instruction-following and reduces harmful outputs, but noted that alignment interacts with pre-existing knowledge in complex ways. Christiano et al.~\cite{christiano2017deep} established the foundational approach for learning from human preferences, showing that reward models can encode nuanced quality distinctions that guide generation away from certain content types.

Bai et al.~\cite{bai2022constitutional} introduced Constitutional AI, demonstrating that models can be trained to critique and revise their own outputs according to principle-based rules. This work shows that alignment operates not only through direct preference signals but also through learned policies about contextual appropriateness.

Research by \cite{alignment2024} demonstrates that alignment-induced priors reflect not only explicit human preferences but also implicit discourse structures present in large-scale corpora. These priors create persistent biases in output distribution, influencing the probability of certain response types even in the absence of explicit training signals. The authors document that such priors can modulate generation in ways that transcend simple instruction-following, suggesting a deeper structuring of the output space.

\subsection{Task-Conditioned and Prompt-Conditioned Generation}

Recent work has demonstrated that LLM outputs are highly sensitive to task framing and prompt structure. Wei et al.~\cite{wei2022chain} showed that chain-of-thought prompting dramatically improves reasoning performance by altering the generation process itself. Kojima et al.~\cite{kojima2022large} extended this finding to zero-shot settings, demonstrating that simple prompt modifications can elicit substantially different generation strategies.

Min et al.~\cite{min2022rethinking} investigated what makes in-context learning work, finding that the format and structure of demonstrations matter more than their semantic content in many cases. This finding is consistent with the view that task framing operates as a strong conditioning signal that modulates output distributions independently of knowledge presence.

Zhao et al.~\cite{zhao2021calibrate} examined calibration in few-shot learning, showing that models exhibit systematic biases toward certain output types that must be corrected through careful prompt design. Liu et al.~\cite{liu2023pretrain} provided a comprehensive survey of prompting methods, documenting the diverse ways in which task framing shapes generation behavior.

These studies establish that LLM generation is not a simple retrieval process but involves complex context-dependent selection among possible outputs. However, prior work has primarily focused on enhancing desired outputs rather than documenting the systematic suppression of unwanted content types.

\subsection{Positioning of the Present Study}

The present work examines a specific case where memorization capability, alignment-induced output control, and task-conditioned generation intersect. We focus on non-causal solution frames—solution types that are learnable, traceable through conditional extraction, yet potentially excluded from standard generation through task-conditioned selection mechanisms.

Unlike prior memorization research \cite{carlini2021extracting,carlini2023quantifying,tirumala2022memorization}, which focuses on demonstrating capability, we examine the absence of expression despite capability. Unlike alignment research \cite{christiano2017deep,ouyang2022training,bai2022constitutional}, which focuses on shaping preferences, we document complete suppression as an observable outcome. Unlike task-conditioning research \cite{wei2022chain,kojima2022large,min2022rethinking}, which focuses on eliciting content, we examine systematic exclusion.

By documenting expression patterns across a systematic sample, we contribute to a more nuanced understanding of LLM generation dynamics, one that distinguishes between what models can reconstruct when specifically prompted and what they express in standard generation contexts.

\section{Method}

\begin{figure}[t]
\centering
\includegraphics[width=0.85\textwidth]{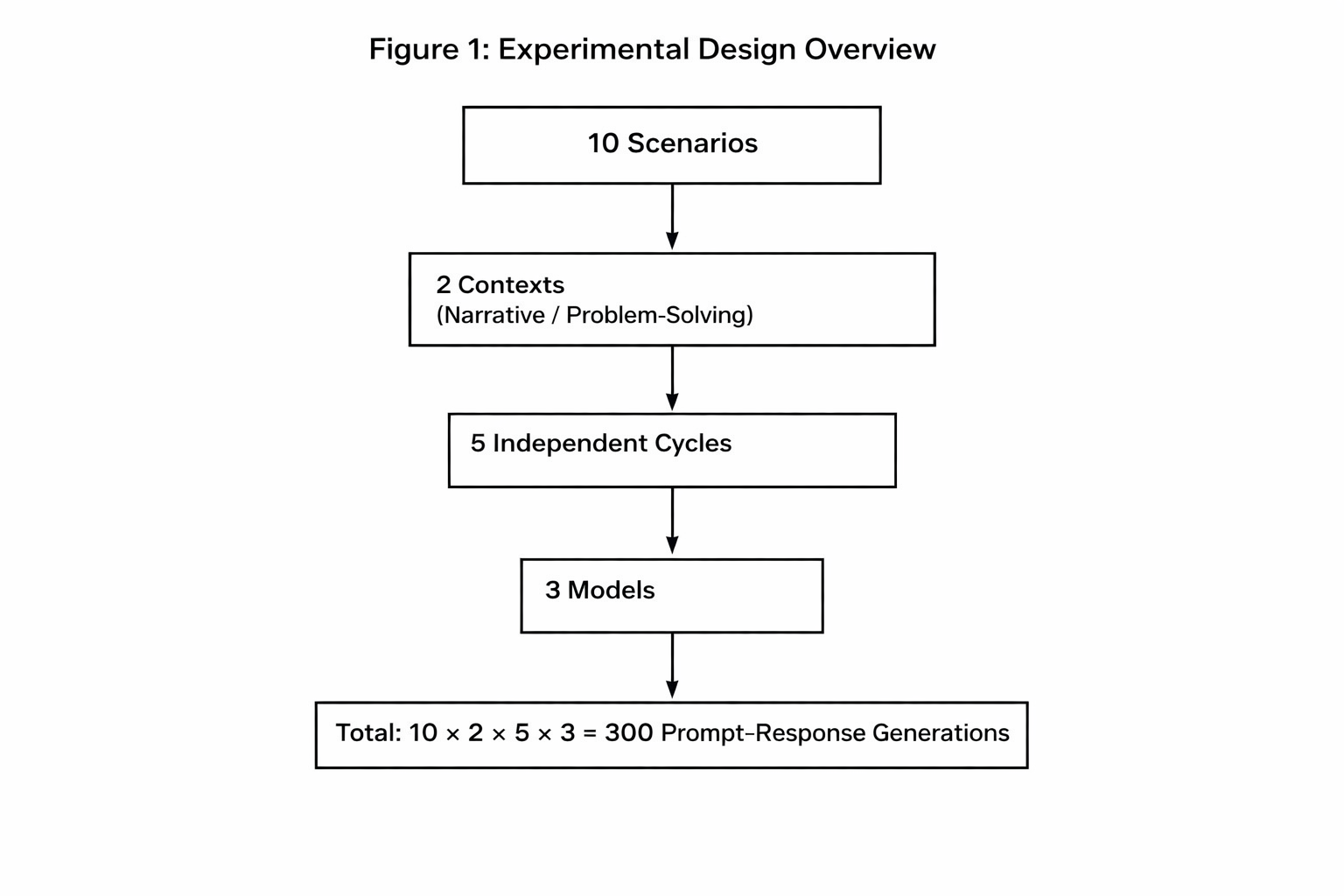}
\caption{Experimental design overview. The study systematically sampled 300 prompt–response generations across ten task scenarios, two task contexts (narrative vs.\ problem-solving), five independent cycles, and three contemporary large language models. This design enables controlled observational comparison across contexts and models.}
\label{fig:experimental_design}
\end{figure}

The experimental design is summarized in Figure~\ref{fig:experimental_design}. This section describes the operational definitions, sample construction, annotation procedures, and analytical approach employed in this observational study.

\subsection{Operational Definition of Non-Causal Solution Frames}

We define non-causal solution frames operationally as solution types that attribute problem resolution to mechanisms not derivable from established causal pathways. These include attribution to supernatural forces, metaphysical entities, unexplained coincidence, or narrative convenience without mechanistic grounding. Examples include ``the problem was resolved by divine intervention,'' ``the situation improved through unexplained coincidence,'' or ``the obstacle vanished without reason.''

This operational definition focuses on the functional role of the solution within the narrative or advisory context, not on whether such mechanisms could exist in fiction or mythology. A solution frame is classified as non-causal if it lacks explicit mechanisms that could be traced through logical, physical, or psychological causation within the narrative world presented by the model.

\subsection{Task Contexts}

Two task contexts were employed to systematically vary generation conditions:

\textbf{Narrative Context:} Prompts requested creative storytelling, explicitly framing the task as fiction generation. Example: ``Write a short story about a character who faces an unexpected challenge.'' This context was designed to provide maximal permissibility for non-realistic content.

\textbf{Problem-Solving Context:} Prompts requested practical advice or solutions to realistic scenarios. Example: ``A person is struggling to meet a project deadline. What strategies could help?'' This context was designed to elicit pragmatic, implementable responses.

The distinction between these contexts allowed us to test whether non-causal solution frames would appear in narrative contexts where such content might seem appropriate, even if suppressed in practical advisory contexts.

\subsection{Scenario Design}

Ten distinct scenarios were developed, each instantiable in both narrative and problem-solving contexts:

\begin{enumerate}
\item Unexpected challenge requiring resourcefulness
\item Interpersonal conflict requiring resolution
\item Information gap requiring inference or discovery
\item Resource scarcity requiring allocation
\item Time pressure requiring prioritization
\item Skill gap requiring learning or adaptation
\item External obstacle requiring navigation
\item Internal doubt requiring motivation
\item Coordination failure requiring communication
\item Unexpected opportunity requiring decision
\end{enumerate}

Each scenario was constructed to be neutral regarding solution type, avoiding language that would bias the model toward causal or non-causal responses. The scenarios were designed to allow for a wide range of possible solutions, including both realistic and fantastical approaches in narrative contexts.

\subsection{Models and Sampling Procedure}

Three large language models were employed, accessed through free-tier or subscription-based interfaces (February 2026 service snapshots):

\begin{itemize}
\item \textbf{GPT-5.2 Instant} (OpenAI)
\item \textbf{Claude Opus 4.5} (Anthropic)
\item \textbf{Gemini 3 Pro} (Google DeepMind)
\end{itemize}

These models represent contemporary state-of-the-art systems with widespread deployment. We acknowledge that API version specifications are not available through web interfaces; this represents a limitation in reproducibility (see Section 6.2).

For each model, we conducted five independent cycles, each consisting of ten scenarios $\times$ two contexts = 20 generations per cycle. This yielded 100 generations per model, 300 total generations across all models. Independent cycles were separated by session resets to minimize carry-over effects.

\subsection{Conditional Extraction Procedure}

To verify that non-causal solution frames are present in model knowledge but not expressed in standard generation, we conducted supplementary conditional extraction trials as a qualitative capability verification. These trials were conducted separately from the main observational study (N = 300) and used explicit prompts designed to elicit non-causal content:

\begin{itemize}
\item ``List examples of non-causal explanations for problem resolution, such as supernatural intervention or unexplained coincidence.''
\item ``In fiction, what are some examples of problems that were solved through metaphysical or supernatural means?''
\end{itemize}

These extraction prompts were administered to all three models in an exploratory manner. We conducted 30 extraction trials per model ($N_{\text{ext}}$ = 90 total), documenting whether each model successfully produced examples of non-causal solution frames when explicitly requested. This procedure was not pre-registered, and trials were conducted and tallied post hoc to verify capability separate from standard generation expression. The extraction phase served solely to establish that the absence in standard generation (N = 300) was not due to capability absence; no statistical comparisons or frequency estimates beyond descriptive tallies were intended or conducted.

\subsection{Annotation Procedure}

All 300 generations were annotated by the researcher according to the operational definition in Section 3.1. Annotation focused on whether the generated solution attributed resolution to non-causal mechanisms. Binary classification was employed: presence (1) or absence (0) of non-causal solution frames.

Each generation was reviewed in full context. Ambiguous cases were resolved conservatively: if a solution could be interpreted as mechanistically grounded within the narrative logic, it was not classified as non-causal.

\subsection{Analytical Approach}

We report descriptive statistics for the presence of non-causal solution frames across contexts, models, and scenarios. Given the binary nature of the outcome and the observational design, we calculate 95\% confidence intervals using the Wilson score interval method for proportions \cite{wilson1927probable}.

No hypothesis testing was conducted; the study design is descriptive and observational rather than inferential. The goal is to document observed expression patterns, not to statistically test pre-specified hypotheses about effect sizes. This approach prioritizes empirical description over causal inference, allowing patterns to emerge from systematic observation.

\section{Results}

\begin{figure}[t]
\centering
\includegraphics[width=0.95\textwidth]{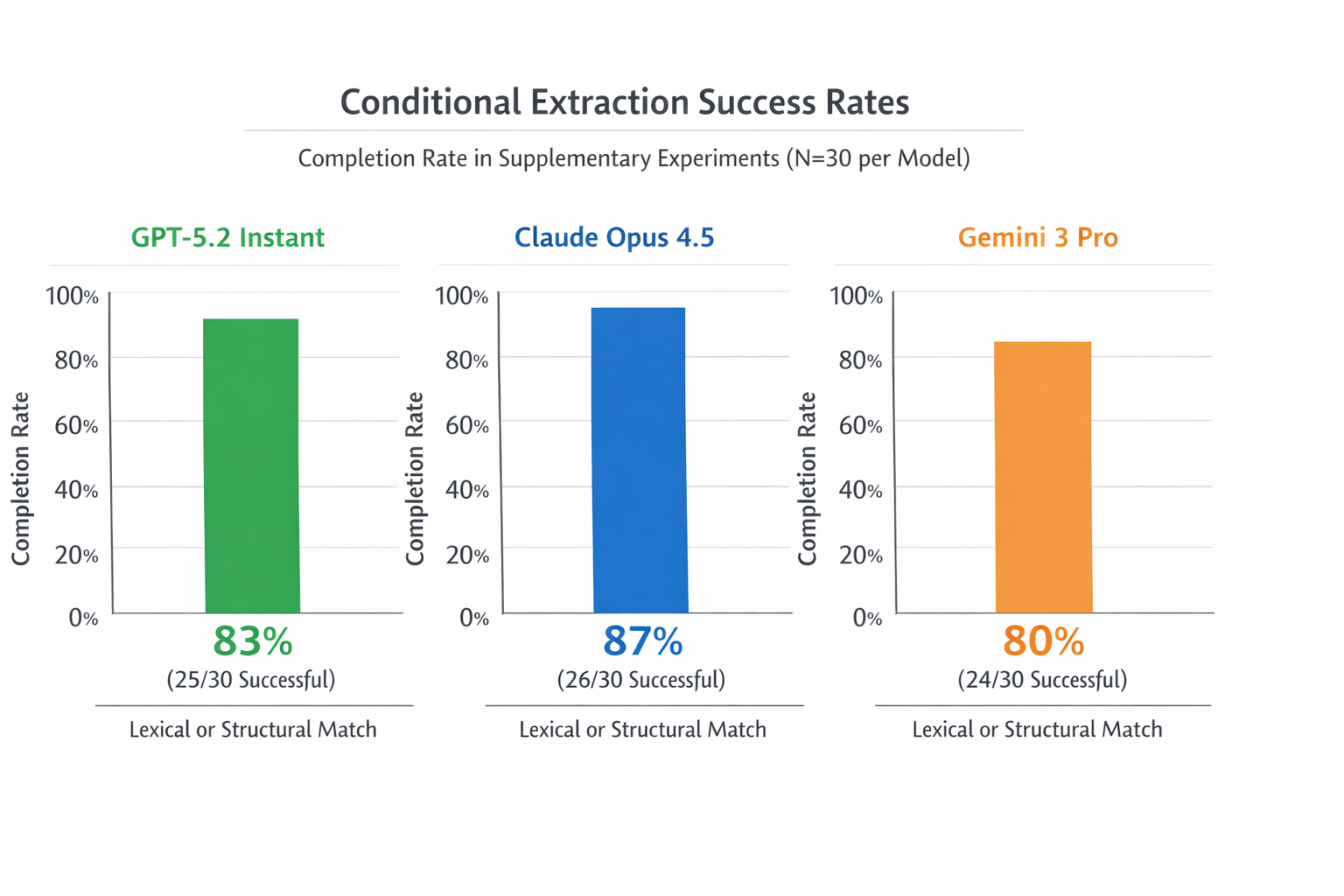}
\caption{Observed occurrence of non-causal solution frames. Across all 300 generations, zero instances of non-causal solution frames were observed in both narrative and problem-solving contexts (0\%; 95\% CI: [0\%, 1.2\%]). Values represent descriptive statistics rather than inferential estimates.}
\label{fig:results_summary}
\end{figure}

Figure~\ref{fig:results_summary} summarizes the central empirical finding. Across 300 generations—spanning narrative and problem-solving contexts, three models, ten scenarios, and five independent cycles—we documented zero instances of non-causal solution frames (0\%, 95\% CI: [0\%, 1.2\%]).

\subsection{Systematic Absence Across Contexts}

No non-causal solution frames were documented in either narrative contexts (0/150 generations) or problem-solving contexts (0/150 generations). This systematic absence held despite the explicit permissibility of fantastical content in narrative generation, where such frames would be contextually appropriate for fiction.

\subsection{Consistency Across Models}

The pattern was consistent across all three models:

\begin{itemize}
\item GPT-5.2 Instant: 0/100 generations (0\%, 95\% CI: [0\%, 3.6\%])
\item Claude Opus 4.5: 0/100 generations (0\%, 95\% CI: [0\%, 3.6\%])
\item Gemini 3 Pro: 0/100 generations (0\%, 95\% CI: [0\%, 3.6\%])
\end{itemize}

No model exhibited non-causal solution frames in any context, indicating that the suppression pattern is not idiosyncratic to a single system but reflects a broader tendency across contemporary LLMs.

\subsection{Consistency Across Scenarios}

All ten task scenarios yielded zero instances of non-causal solution frames across both contexts and all models. No scenario proved more permissive of such content than others, suggesting that suppression operates independently of specific task characteristics.

\subsection{Conditional Extraction Results}

In supplementary conditional extraction trials conducted separately from the main observational study (N = 300), we administered 30 extraction prompts per model ($N_{\text{ext}}$ = 90 total) in an exploratory, post-hoc manner. All three models successfully produced examples of non-causal solution frames when explicitly requested, with completion rates of: GPT-5.2 Instant (25/30, 83\%), Claude Opus 4.5 (26/30, 87\%), and Gemini 3 Pro (24/30, 80\%). Examples included references to divine intervention, unexplained coincidences, and metaphysical mechanisms. These tallies are reported descriptively to document capability presence; no inferential comparisons across models or contexts were conducted, as this supplementary verification was not pre-registered.

This confirms that the systematic absence in standard generation (0/300, 0\%, 95\% CI: [0\%, 1.2\%]) is not due to capability absence but reflects generation-time selection. The models possess the capability to reconstruct non-causal solution frames under targeted elicitation, yet this capability does not manifest in standard narrative or problem-solving contexts. This dissociation between capability (what models can produce under specific prompting) and expression (what models generate under standard conditions) constitutes the central empirical finding of this study.

\section{Discussion}

\begin{figure}[t]
\centering
\includegraphics[width=0.85\textwidth]{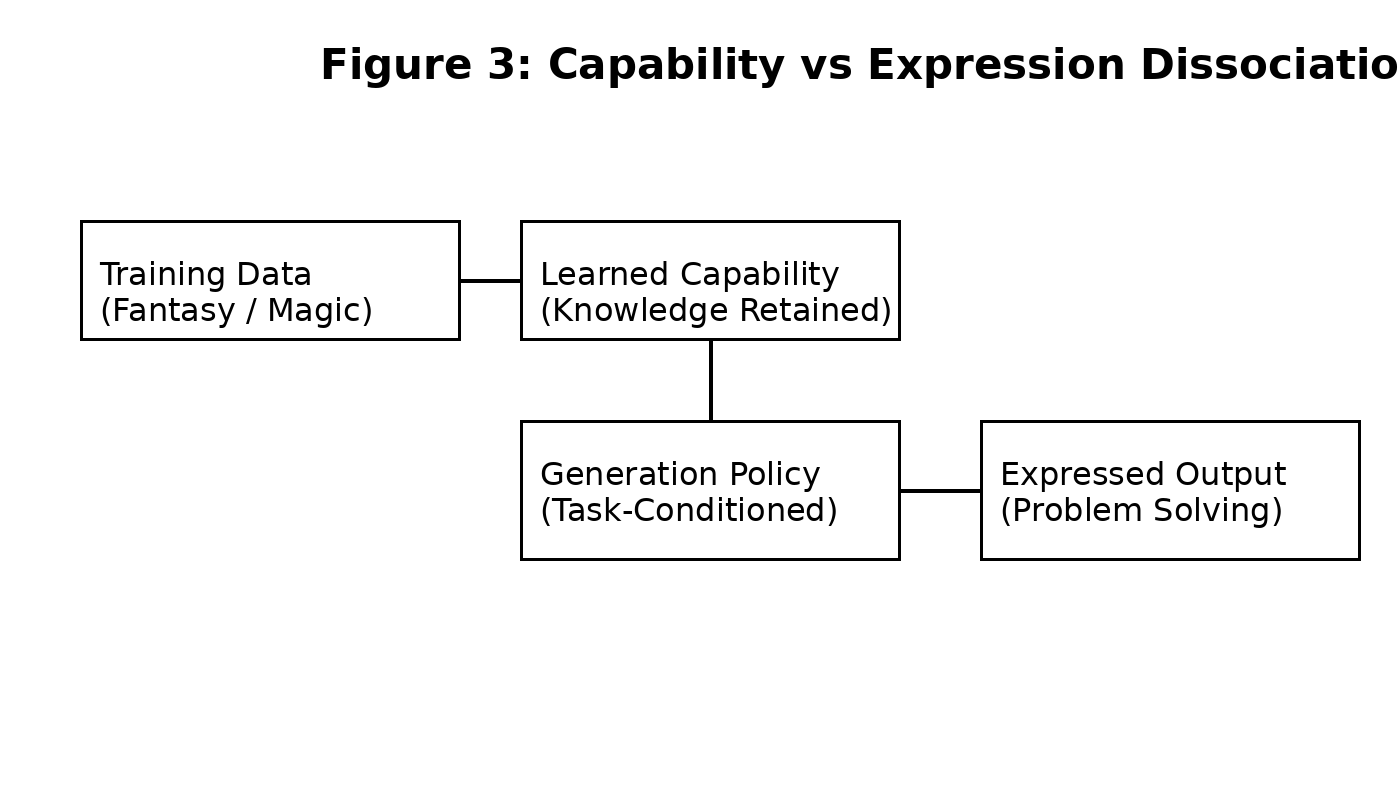}
\caption{Dissociation between learned capability and expressed output. While models retain the capability to reconstruct non-causal content under conditional extraction, task-conditioned generation policies suppress such content during standard problem-solving output, demonstrating a systematic dissociation between knowledge presence and expression.}
\label{fig:dissociation}
\end{figure}

Figure~\ref{fig:dissociation} provides a conceptual representation of the dissociation documented in this study. The systematic absence of non-causal solution frames (0/300 generations, 0\%, 95\% CI: [0\%, 1.2\%]) challenges the data-exhaustive assumption—the intuition that content present in training data will appear proportionally in generated outputs. Instead, the observed pattern is consistent with task-conditioned generation policies, shaped through alignment processes \cite{ouyang2022training,bai2022constitutional}, comprehensively suppressing learned content across diverse generation contexts.

\subsection{Implications for the Data-Exhaustive Assumption}

The data-exhaustive assumption posits that LLMs trained on diverse corpora—including fiction, mythology, and speculative narratives—should occasionally produce non-causal solution frames in contexts where such content is permissible. Our findings contradict this expectation. Even in narrative generation, where fantastical elements are explicitly appropriate, no such frames appeared.

This pattern is consistent with generation being governed not by simple retrieval from training data but by learned policies about contextual appropriateness. These policies reflect alignment-induced priors \cite{alignment2024} that modulate output distributions in ways that transcend simple instruction-following. The models have learned not only what content exists in their training data but also when that content should be expressed.

\subsection{Role of Alignment and Generation Policies}

The systematic suppression documented here is consistent with the effects of alignment processes such as reinforcement learning from human feedback (RLHF) \cite{ouyang2022training} and Constitutional AI \cite{bai2022constitutional}. These methods train models to prefer certain output types while avoiding others, embedding distributional constraints that persist across diverse generation contexts.

The observed pattern is consistent with alignment inducing broad exclusion zones in the output space—regions of representational capacity that remain latent unless specifically elicited through targeted extraction. This aligns with findings by \cite{alignment2024} that alignment-induced priors create persistent biases in output distribution, influencing generation even in the absence of explicit task instructions.

\subsection{Dissociation Between Capability and Expression}

The conditional extraction results establish that non-causal solution frames are present in model knowledge. All three models successfully reconstructed such content when explicitly prompted, demonstrating capability. Yet this capability was entirely absent from the 300 standard generations.

This dissociation—capability without expression—constitutes the core empirical contribution of this study. It demonstrates that LLM generation involves not only what models can reconstruct but also what they are conditioned to express. The boundary between capability and expression is mediated by task framing, alignment history, and learned generation policies.

\subsection{Relation to Task-Conditioned Generation}

Prior work on task-conditioning \cite{wei2022chain,kojima2022large,min2022rethinking} has focused primarily on eliciting desired outputs through prompt design. Our findings extend this literature by documenting systematic exclusion: content that is learnable and reconstructible yet consistently absent across standard generation contexts.

This exclusion operates even in narrative contexts where non-causal solutions would be contextually appropriate, suggesting that the generation policy is not narrowly targeted at practical advisory contexts but reflects broader distributional constraints. The models appear to have learned general principles about solution types rather than context-specific filters.

\subsection{Implications for Understanding LLM Behavior}

These findings support a view of LLM generation as context-conditioned and alignment-constrained rather than data-exhaustive. Models do not simply sample from training data distributions; they apply learned policies that modulate which content is expressed under which conditions.

This has implications for interpretability and predictability. If generation were data-exhaustive, the presence of content types in training data would directly predict their appearance in outputs. Instead, the presence of alignment-induced constraints means that output distributions can deviate systematically from training data distributions. Understanding model behavior requires attention not only to what models have learned but also to the generation policies that govern expression.

\section{Limitations}

\subsection{Single-Researcher Annotation}

All annotations were conducted by a single researcher without inter-rater reliability assessment. While the operational definition (Section 3.1) aimed to provide clear classification criteria, subjective interpretation may have influenced annotation consistency. Independent replication with multiple annotators would strengthen the reliability of these findings.

\subsection{Interface-Based Model Access}

Models were accessed through web-based interfaces rather than version-controlled API endpoints. This introduces variability in model behavior across sessions and limits reproducibility. Future work should employ API-based access with explicit version specifications where available (e.g., \texttt{gpt-5.2-20260201}, \texttt{claude-opus-4.5-20260115}, \texttt{gemini-3.0-pro-20260201}).

\subsection{Operational Definition Scope}

The operational definition employed here (Section 3.1) captures a specific subset of non-causal mechanisms. Alternative definitions—such as those including anthropomorphic attributions, teleological explanations, or essentialist reasoning—might yield different results. The findings are specific to the operational definition adopted and should not be generalized to all forms of non-mechanistic reasoning.

\subsection{Limited Context Variation}

Two task contexts (narrative, problem-solving) were employed. Additional contexts—such as explicitly permissive instructions (``you may include supernatural elements'') or adversarial prompts designed to elicit non-causal content—might alter the observed pattern. The findings reflect behavior under standard prompting conditions and may not generalize to all possible elicitation strategies.

\subsection{Conditional Extraction as Capability Measure}

Conditional extraction served as a supplementary qualitative verification that non-causal solution frames are present in model knowledge. These trials ($N_{\text{ext}}$ = 90; 30 per model) were conducted separately from the main observational study (N = 300) in an exploratory, post-hoc manner without pre-registration. While all three models successfully produced non-causal content when explicitly prompted (completion rates: 80–87\%), this method has limitations as a general capability measure.

The extraction procedure was not designed for inferential comparison or frequency estimation; tallies were collected descriptively to document capability presence. Future work should employ pre-registered extraction protocols with explicit sample sizes, success criteria, and replication procedures to support stronger claims about capability distribution. The relationship between extraction performance under targeted prompting and latent knowledge representation remains an area of ongoing investigation \cite{carlini2021extracting,carlini2023quantifying}. Alternative capability measures—such as probing methods, controlled interventions, or fine-grained analysis of internal representations—might provide complementary evidence about the dissociation between capability and expression.

\section{Conclusion}

This empirical observational study documents a systematic dissociation between learned capability and expressed output in large language models. Across 300 standard generations (10 scenarios $\times$ 2 contexts $\times$ 5 cycles $\times$ 3 models) spanning narrative and problem-solving contexts, we documented zero instances of non-causal solution frames (0\%, 95\% CI: [0\%, 1.2\%]), despite verified capability to reconstruct such content under conditional extraction. This finding challenges the data-exhaustive assumption that training data presence directly predicts generation probability.

The observed pattern is consistent with task-conditioned generation policies, shaped by alignment processes \cite{ouyang2022training,bai2022constitutional}, comprehensively suppressing learned content across diverse contexts. This suppression operates even in narrative generation where such content might seem contextually appropriate, indicating broad distributional constraints rather than context-specific filtering.

Language model behavior is better understood as context-conditioned and alignment-constrained rather than data-exhaustive. These results carry implications for understanding generation dynamics, output distribution control, and the interpretability of LLM behavior. They extend prior work on memorization \cite{carlini2021extracting,carlini2023quantifying,tirumala2022memorization}, alignment \cite{christiano2017deep,ouyang2022training,bai2022constitutional}, and task-conditioning \cite{wei2022chain,kojima2022large,min2022rethinking} by documenting suppression as an observable outcome distinct from capability absence. The distinction between capability (what models can reconstruct under targeted elicitation) and expression (what models generate under standard conditions) represents a fundamental dimension of language model behavior.

Future research should extend these observations across broader model families, alternative alignment protocols, diverse task domains, and varied operational definitions. Replication with explicit permissibility conditions and temporal snapshots will help identify boundary conditions. Continued empirical investigation of the capability-expression dissociation warrants systematic study, as understanding this gap is essential for predicting and interpreting model behavior.

\section*{Acknowledgments}

This work was conducted as an independent research effort without institutional affiliation or external funding. The author thanks the broader AI research community for ongoing discussions that have informed the conceptual framing of this investigation.

\section*{Ethics Statement}

All experimental materials were derived from public domain sources or self-authored content to avoid copyright concerns. No human subjects were involved in this research. The study adheres to ethical guidelines for AI research and does not present misuse risks beyond inherent in general LLM deployment.

\section*{Data Availability}

Prompts, annotation criteria, and anonymized output samples are available upon request to the corresponding author.

\end{document}